\documentclass[letterpaper]{article} 
\usepackage{aaai2027}  
\usepackage[hyphens]{url}  
\usepackage{graphicx} 
\usepackage{natbib}  
\usepackage{caption} 
\usepackage{algorithm}
\usepackage{algorithmic}

\usepackage{newfloat}
\usepackage{listings}
\DeclareCaptionStyle{ruled}{labelfont=normalfont,labelsep=colon,strut=off} 
\floatstyle{ruled}
\newfloat{listing}{tb}{lst}{}
\floatname{listing}{Listing}

\usepackage{booktabs}

\usepackage{cleveref}
\usepackage{comment} 
\usepackage{enumitem} 
\usepackage{multirow}
\usepackage{subcaption} 
\usepackage{soul} 

\title{Pre-Train To Gain: In-Domain Self-Supervised Pre-Training For Robust Learning}
\author{
    David Szczeinca\thanks{\hspace{1mm}We acknowledge the support of the Natural Sciences and Engineering Research Council of Canada (NSERC) through the NSERC Discovery program and the NSERC CGRS-M program.},
    Nicholas Pellegrino,
    Paul Fieguth,
}
\affiliations{
    Systems Design Engineering\\
    University of Waterloo\\
    Waterloo, Ontario, Canada \\
    \textit{david.szczecina@uwaterloo.ca}
}

\begin{document}

\maketitle

\begin{abstract}
Training deep networks with noisy labels leads to poor generalization and degraded accuracy due to overfitting to label noise. Existing approaches for learning with noisy labels often rely on the availability of a clean subset of data. 
By pre-training a feature extractor on the target dataset without labels using in-domain self-supervised learning (SSL), followed by standard supervised training on the same noisy dataset, we can train a more noise robust model without requiring a subset with clean labels. 
We evaluate both contrastive and non-contrastive SSL pre-training methods across datasets with synthetic and real-world label noise, demonstrating the broad applicability of our approach across large-scale datasets, diverse downstream tasks, and model architectures.
Across all noise rates, in-domain self-supervised pre-training consistently improves classification accuracy and downstream label-error detection (F1 and Balanced Accuracy) compared with supervised training from scratch. The performance gap widens as the noise rate increases, demonstrating improved robustness. 
Notably, our approach achieves comparable results to ImageNet and DinoV2 pre-trained models at low noise levels, while substantially outperforming them under high noise conditions.
\end{abstract}

\begin{links}
    \link{Code}{https://anonymous.4open.science/r/SSL_Pre-Train}
\end{links}

\section{Introduction}
\label{sec:intro}

Deep neural networks have achieved remarkable success across a wide range of supervised learning tasks \citep{krizhevsky2012imagnet,he2016deep}. However, their performance depends heavily on the quality of labeled data. In real-world settings, datasets often contain noisy labels due to human annotation errors, automated labeling processes, or ambiguous data, where $\eta\in[0,1]$ denotes the fraction of labels that are incorrect \citep{zhang2018generalized,northcutt2021confident}. 
Training directly on such noisy labels typically leads to poor generalization, as deep networks are prone to memorizing the noise \citep{song2022learning,zhang2016understanding,pleiss2020identifying}, especially in the absence of explicit noise-handling mechanisms.
A common strategy to mitigate this issue is to identify and correct noisy labels using robust loss functions or noise-aware training strategies \citep{Li2020dividemix, han2018co-sampling, wei2020combating, zheltonozhskii2021c2d}. Many of these methods, however, assume access to a small subset of clean labels ($\eta=0$) or require prior knowledge about the noise distribution \citep{Li2020dividemix, Liu2020early, han2018co-sampling}, assumptions that are often unrealistic in practical scenarios.

\begin{figure}[t]
    \centering
    \includegraphics[width=\columnwidth]{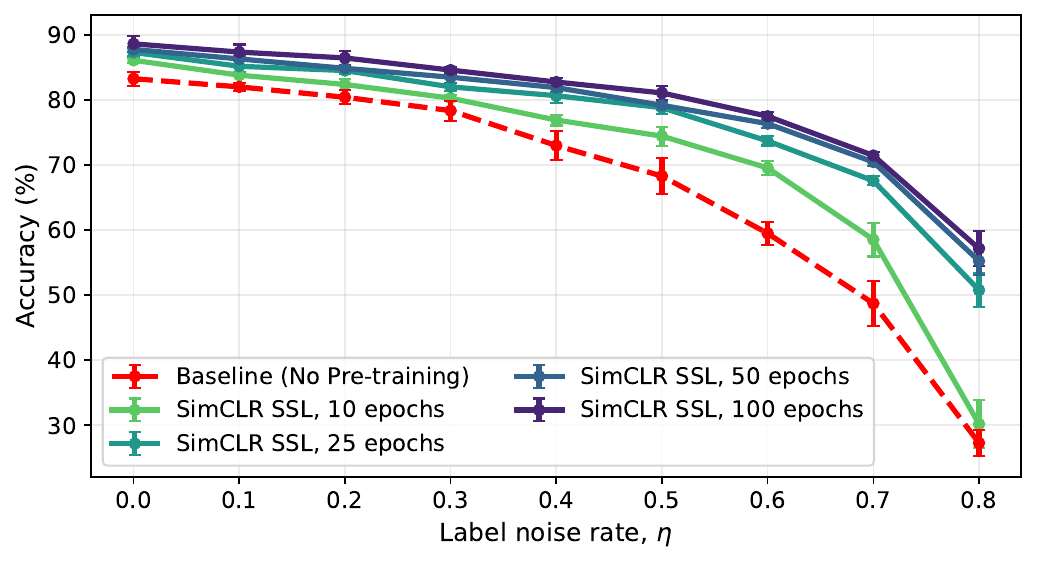}
    \caption{Top-1 Classification accuracy across varying label noise rates, for increasing durations of self-supervised pre-training on CIFAR-10 using SimCLR. As the number of SSL pre-training epochs increases, downstream accuracy consistently improves across all corruption levels. 
    }
    \label{simclr_acc_cifar10}
    \vspace{-2mm}
\end{figure}

Self-supervised learning (SSL) offers a promising alternative by allowing feature representations to be learned without using labels at all \citep{chen2020simclr}. 
Traditionally, SSL is performed on large external datasets to obtain transferable representations. In contrast, this work investigates in-domain SSL, where pre-training is performed directly on the target training dataset rather than an external dataset, prior to supervised learning under label noise. By using SSL, feature extractors can be trained without labels, thereby avoiding any negative influences of mislabelled data.

In this work, we investigate whether self-supervised pre-training can enhance a model's noise robustness when training on noisy-labelled datasets, without relying on any clean subset of data. 
We compare supervised training from scratch with a two-stage approach in which the model is first pre-trained using in-domain self-supervised learning before supervised fine-tuning on the same noisy dataset.
As seen by results in \Cref{simclr_acc_cifar10}, self-supervised pre-training consistently improves model robustness. Specifically, we observe higher test accuracy, evaluated on clean test sets, and improved label error detection capability when predicting on noisy test sets.

To validate these claims, we perform a progressively broader series of experiments. We first establish that in-domain SSL improves classification accuracy and label-error detection under synthetic label noise. We then demonstrate that these gains extend to real-world annotation errors, persist under longer supervised training, and arise from improved feature representations. Finally, we examine the generality of the approach by evaluating larger-scale datasets, semantic segmentation, and Vision Transformer architectures \citep{dosovitskiy2021imageworth} using multiple modern SSL objectives. Across all of these settings, in-domain SSL consistently improves robustness without modifying the downstream supervised learning algorithm.
This study demonstrates that self-supervised pre-training on the target dataset can serve as a simple yet powerful method for improving model robustness to label errors. Furthermore, we show that in-domain SSL can outperform strong generic pre-trained initializations and that combining generic pre-training with continued in-domain SSL yields further performance improvements.

\section{Background}

\paragraph{Learning with Noisy Labels}
Supervised learning assumes access to correctly labelled data, but in real-world scenarios, label noise is common~\citep{northcutt2021pervasive}. 
Training neural networks directly on noisy data often results in memorization~of incorrect labels, leading to poor generalization and overfitting~\citep{song2022learning,zhang2016understanding,pleiss2020identifying}.
Numerous strategies have been proposed to address training with label noise, including: 
robust loss functions that reduce the model's sensitivity to incorrect labels \citep{pellegrino2024loss, zhang2018generalized, ye2023active},
sample re-weighting to prioritize clean samples \citep{han2018co-sampling, jiang2018mentornet}.
semi-supervised and weakly supervised approaches that leverage a clean subset to guide training \citep{Li2020dividemix, Liu2020early}.
A recurring limitation across many of these approaches is the reliance on a small subset of data with clean labels, which may not be feasible in large-scale or low-resource settings \citep{northcutt2021pervasive}.

\paragraph{Self-Supervised Learning (SSL)}
Self-supervised learning has emerged as a powerful method for representation learning without requiring any labels \citep{chen2020simclr, zbontar2021barlow, grill2020byol}. In computer vision, SSL methods typically use data augmentations and auxiliary pretext tasks to learn meaningful image embeddings. Two widely used approaches are 



\begin{itemize}[nosep] 
    \item \textbf{SimCLR:} (Simple Framework for Contrastive Learning) A contrastive learning method that trains encoders to bring augmented views of the same image closer in representation space while pushing apart views from different images \citep{chen2020simclr}.
    \item\textbf{Barlow Twins:} A redundancy-reduction based method that encourages representations of augmented views to be similar while reducing redundancy between the components of the embeddings \citep{zbontar2021barlow}.
\end{itemize}

These methods have been shown to learn generalizable and semantically rich features, even without labels. More recently, SSL has been successfully extended to Vision Transformer architectures through methods such as DINO, DINOv2, iBOT, MAE, MSN, SimMIM, and VICReg \citep{caron2021dino, oquab2024dinov2, zhou2022ibot, he2022mae, assran2023msn, xie2022simmim, bardes2022vicreg}. When used for pre-training, these representations provide improved model initialization and stronger downstream performance.

\paragraph{Label Error Detection}
Identifying mislabelled instances is a critical step in improving training on noisy datasets. Frameworks for detecting label errors typically require well-trained and well-generalized models to perform effectively. Several techniques have been proposed for this purpose, ranging from training dynamics based methods \citep{pleiss2020identifying} to confidence-based filtering. Confident Learning~\citep{northcutt2021confident} is a prominent framework that estimates the probability that each example’s given label is incorrect. It does so using the predicted probabilities and the confusion matrix estimated from the model’s outputs.

\paragraph{Related Works}
Contrast to Divide (C2D) \citep{zheltonozhskii2021c2d} identified a key limitation in learning with noisy labels, the “warm-up obstacle”, where models quickly begin memorizing incorrect labels during initial supervised training. C2D addresses this by introducing a two-stage framework: a self-supervised contrastive pre-training phase followed by integration with existing learning-with-noisy-labels (LNL) algorithms such as DivideMix \citep{Li2020dividemix} or ELR+ \citep{Liu2020early}. This pre-training step provides noise-invariant features that improve label separation and classification accuracy.
In contrast to C2D, our work isolates the effect of self-supervised pre-training itself, without coupling it to any LNL algorithm or specialized loss. Rather than improving a specific warm-up mechanism, we show that stand-alone SSL pre-training on the same noisy dataset consistently enhances downstream performance for both accuracy and the model's ability to detect label-errors. 

More recently, LeJEPA~\citep{balestriero2025lejepa} proposed a self-supervised learning framework specifically designed for scalable in-domain pre-training, demonstrating that training directly on target-domain data can outperform transfer learning from large external datasets such as DINOv2 on domain-specific tasks. Instead, we investigate whether in-domain self-supervised pre-training itself is the primary source of these gains. By evaluating a diverse range of existing SSL methods, including contrastive, non-contrastive, masked image modeling, and joint-embedding approaches, across both convolutional and Vision Transformer architectures, we show that the benefits of in-domain pre-training are not unique to a particular SSL algorithm, but generalize across modern SSL algorthims.

\section{Method}

We consider supervised image classification in the presence of label noise. Given a training dataset $\mathcal{D}={(x_i,\tilde y_i)}_{i=1}^N$, where a fraction $\eta$ of labels are corrupted, the objective is to learn a classifier that generalizes well to clean test data without requiring clean training labels, prior knowledge of the noise process, or specialized learning-with-noisy-labels algorithms. Our goal is to determine whether in-domain self-supervised pre-training alone provides a more robust initialization for subsequent supervised learning on noisy labels.

To evaluate the effectiveness of self-supervised pre-training in enhancing robustness to label noise, we compare a baseline standard supervised training method against a two-stage training pipeline that incorporates SSL pre-training.

\begin{itemize}[nosep] 
    \item \textbf{Baseline Supervised Method}: The model is trained from scratch on the noisy dataset for 10 epochs. 
    \item \textbf{Self-Supervised + Supervised Fine-tuning}: An identical model is first pre-trained using an SSL method, for a specified number of epochs, before being fine-tuned via 10-epoch supervised training on the noisy dataset.
\end{itemize}



\paragraph{Self-Supervised Methods}
Throughout this paper, \emph{SSL pre-training} refers to the proposed two-stage pipeline, where an encoder is first pre-trained on the target training images without labels and subsequently fine-tuned using supervised learning on the noisy dataset. SimCLR~\citep{chen2020simclr} is used as the default SSL method due to its simplicity and strong empirical performance, following the standard augmentation scheme of random cropping, colour jitter, and Gaussian blur. To demonstrate that the observed robustness improvements are not specific to a single SSL objective, we additionally evaluate the non-contrastive Barlow Twins SSL method \citep{zbontar2021barlow}. Furthermore, to verify that the proposed pipeline generalizes beyond convolutional architectures, we extend our evaluation to Vision Transformers using several modern SSL methods, including DINO, \mbox{DINOv2}, iBOT, MAE, MSN, SimMIM, and VICReg.

\begin{table*}[t]
\centering
\caption{Comparison of Classification Accuracy (Acc; \%), F1-score (F1), and Balanced Accuracy (BA) under varying noise rates $\eta$ on CIFAR-10 and CIFAR-100. Results are statistically significant ($t$-test, $p<0.05$), and best scores are bolded. Self-supervised pre-training for 100 epochs with either SimCLR or Barlow Twins consistently improves performance under label noise.}

\label{tab:Main_results}
\resizebox{\textwidth}{!}{
\begin{tabular}{llc|ccc|ccc|ccc|ccc}
\toprule
\multirow{2}{*}{\textbf{Dataset}} &
\multirow{2}{*}{\textbf{Method}} &
\multicolumn{1}{c|}{$\eta=0.0$} &
\multicolumn{3}{c|}{$\eta=0.2$} &
\multicolumn{3}{c|}{$\eta=0.4$} &
\multicolumn{3}{c|}{$\eta=0.6$} &
\multicolumn{3}{c}{$\eta=0.8$} \\
\cmidrule(lr){3-15}
 &  & Acc & Acc & F1 & BA & Acc & F1 & BA & Acc & F1 & BA & Acc & F1 & BA \\
\midrule
\multirow{3}{*}{CIFAR-10} & Baseline & 83.27 & 80.42 & 0.80 & 0.91 & 73.00 & 0.85 & 0.87 & 59.50 & 0.83 & 0.79 & 27.28 & 0.73 & 0.61 \\
 & SimCLR & \textbf{88.63} & \textbf{86.46} & \textbf{0.84} & \textbf{0.93} & \textbf{82.76} & \textbf{0.90} & \textbf{0.92} & \textbf{77.50} & \textbf{0.91} & \textbf{0.88} & \textbf{57.16} & \textbf{0.76} & \textbf{0.73} \\
 & Barlow Twins & 87.95 & 85.09 & 0.83 & 0.92 & 81.74 & 0.89 & 0.91 & 74.89 & 0.90 & 0.87 & 51.12 & 0.75 & 0.71 \\
\midrule
\multirow{3}{*}{CIFAR-100} & Baseline & 56.21 & 50.03 & 0.57 & 0.78 & 41.40 & 0.72 & 0.75 & 27.42 & 0.77 & 0.69 & 9.74 & 0.68 & 0.61 \\
 & SimCLR & 59.41 & 55.17 & 0.59 & 0.79 & 50.43 & 0.75 & 0.78 & 41.79 & 0.82 & 0.75 & 22.95 & 0.77 & 0.67 \\
 & Barlow Twins & \textbf{62.76} & \textbf{58.07} & \textbf{0.63} & \textbf{0.81} & \textbf{53.05} & \textbf{0.76} & \textbf{0.80} & \textbf{45.31} & \textbf{0.83} & \textbf{0.77} & \textbf{30.89} & \textbf{0.82} & \textbf{0.71} \\
\bottomrule
\end{tabular}
}
\end{table*}

\paragraph{Datasets and Artificial Corruption}
Synthetic uniform label noise is injected into standard benchmark datasets following the same method used in ~\citet{pellegrino2023effects}. For a given corruption rate $\eta \in [0,1]$, a fraction $\eta$ of the training labels are selected uniformly at random and reassigned to one of the remaining (incorrect) class labels with equal probability. Two benchmark image classification datasets are used, CIFAR-10 and CIFAR-100 \citep{krizhevsky2009learning}, under the assumption that their existing label error rates~\citep{northcutt2021pervasive} are negligible compared to the induced corruption,~$\eta$. 
Real-world noise is additionally evaluated using the CIFAR-N datasets~\citep{wei2022learning}, which contain human annotation errors rather than synthetically generated labels, providing a more realistic evaluation of noisy-label robustness. CIFAR-100N has a noise rate of $\eta \approx 0.4$, and CIFAR-10N provides two noisy label sets with $\eta \approx 0.1$ and $\eta \approx 0.4$. We further evaluate the proposed pipeline on Clothing-1M ($\eta \approx 0.4$), PASCAL VOC 2012, and EuroSAT to assess robustness on a large-scale real-world noisy dataset, semantic segmentation, and Vision Transformer architectures, respectively.


\paragraph{Label Error Detection}
To assess whether the learned representations improve identification of mislabeled samples, we apply Confident Learning~\citep{northcutt2021confident} to the final trained model. Since Confident Learning identifies label errors from the model's predicted class probabilities, improved label-error detection provides an indirect measure of representation quality and robustness to label noise. After supervised fine-tuning is complete, the model is evaluated on the corrupted test set to obtain class probability predictions, which Confident Learning uses to estimate which samples are likely to be incorrectly labeled. We then compute the Balanced Accuracy and F1 score of these predicted label errors.
Strong label-error detection indicates that the model has learned representations that reflect the underlying class structure rather than memorizing corrupted labels. Consequently, these metrics provide a complementary measure of robustness beyond classification accuracy alone.

\paragraph{Model Training}
To ensure fairness, models are trained only for the number of epochs at which point overfitting begins on the original uncorrupted datasets. A precursor experiment was performed for both uncorrupted datasets whereby the crossover point between training and validation loss is measured. In both cases, overfitting was found to occur starting at roughly 10 epochs. Unless otherwise stated, experiments use a ResNet-18 backbone and the Adam optimizer~\citep{kingma2014adam}. Complete implementation details, including architectures, optimizers, hyperparameters, and training settings, are provided in Appendix \ref{sec:appendix}.

\section{Results}
The evaluation is organized to progressively validate the proposed approach. We first establish the effectiveness of in-domain SSL under controlled synthetic label noise before examining label-error detection and real-world noisy datasets. We then investigate the mechanisms behind the observed improvements through training-duration and representation analyses. Finally, we evaluate the generality of the approach across larger datasets, semantic segmentation, and Vision Transformer architectures. In all experiments, \textit{Baseline} denotes training from random initialization without SSL or external pre-trained weights. Unless otherwise stated, all experiments were repeated over five random seeds, and results are reported as the mean ± one standard deviation.

\paragraph{Main experiment}
In~\Cref{tab:Main_results}, we compare supervised training from random initialization with models using SSL pre-training via either SimCLR or Barlow Twins. Pre-training was performed for 100 epochs, and all results are averaged over five seeds.
Across both CIFAR-10 and CIFAR-100, SSL pre-training consistently outperforms the supervised baseline at every noise level. Both SimCLR and Barlow Twins improve clean test accuracy, with the performance gap widening as the corruption rate increases, demonstrating greater robustness to label noise. Although SimCLR achieves the highest accuracy on CIFAR-10 and Barlow Twins on CIFAR-100, these differences are small compared to the consistent gains from either SSL pre-training method.
Confident Learning is used to detect label errors on the noisy training set, with F1 score and Balanced Accuracy reported. Both SSL methods consistently improve label-error detection over the baseline, increasing F1 and Balanced Accuracy by 4-7 percentage points on average across both datasets.

\begin{figure}[tb]
    \centering
    \includegraphics[width=\columnwidth]{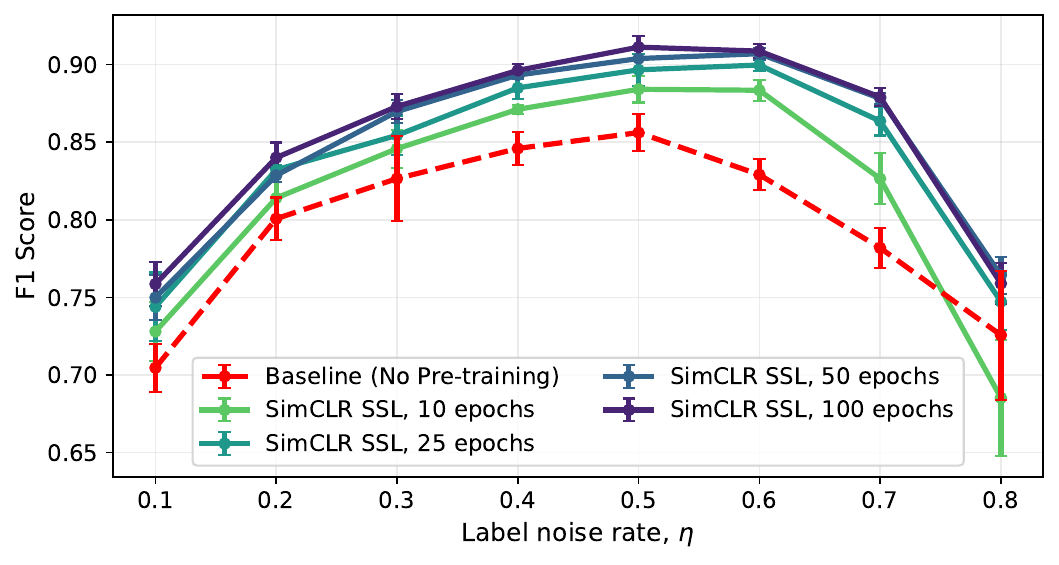}
    \caption{Effect of SimCLR SSL pre-training duration on label error detection (F1-score) on CIFAR-10. Performance improves rapidly with additional pre-training, with gains plateauing after approximately 50 epochs.}
    \label{simclr_f1_cifar10}
\end{figure}

\paragraph{Pre-train Duration Variation} 
We examine how the duration of SSL pre-training influences downstream robustness by evaluating models pretrained for different numbers of epochs. As shown in \Cref{simclr_acc_cifar10,simclr_f1_cifar10}, both classification accuracy and F1 score improve substantially after just 10 epochs of SimCLR pre-training, with gains beginning to plateau after approximately 50 epochs. This trend is consistent across both CIFAR-10 and CIFAR-100, and across both SSL methods evaluated. Balanced Accuracy exhibits the same pattern, indicating that early-stage pre-training provides most of the robustness benefit, with diminishing returns beyond moderate training durations. During this early stage, the encoder learns clean, noise-agnostic features before any corrupted labels can bias the representation.

\paragraph{Real-World Noise} 
To evaluate performance under real-world, non-uniform label errors, we test on the CIFAR-10N and \mbox{CIFAR-100N} datasets, which contain multiple alternative label sets with human annotation noise.
Based on the optimal pre-training duration identified in \Cref{simclr_acc_cifar10}, we pre-train for 50 epochs, capturing most downstream gains while avoiding diminishing returns. Pre-training improves classification accuracy by 4-6\% across all datasets in \Cref{tab:CIFAR-N}. The consistent gains across both CIFAR-10N and CIFAR-100N indicate that the benefits of in-domain SSL extend beyond synthetic label corruption to naturally \mbox{occurring annotation errors.}

\begin{table}[ht]
\centering
\caption{Performance comparison on real-world noisy datasets CIFAR-10N and CIFAR-100N.
Accuracy, F1-score, and Balanced Accuracy are shown for baseline and SimCLR-pretrained models.}
\label{tab:CIFAR-N}
\resizebox{\columnwidth}{!}{
\begin{tabular}{c|c|c|cc}
\toprule
Dataset & $\eta$ & Metric & Baseline & SimCLR (SSL) \\
\midrule
\multirow{3}{*}{CIFAR-10N} & \multirow{3}{*}{$\approx 0.1$}
& Acc. & 81.14 $\pm$ 0.68 & \textbf{85.54 $\pm$ 0.79} \\
& & F1 & 53.70 $\pm$ 2.37 & \textbf{59.69 $\pm$ 1.94} \\
& & BA & 74.48 $\pm$ 1.59 & \textbf{76.40 $\pm$ 0.74} \\
\midrule
\multirow{3}{*}{CIFAR-10N} & \multirow{3}{*}{$\approx 0.4$}
& Acc. & 72.73 $\pm$ 1.60 & \textbf{78.71 $\pm$ 1.02} \\
& & F1 & 77.46 $\pm$ 1.13 & \textbf{80.74 $\pm$ 0.87} \\
& & BA & 81.18 $\pm$ 0.75 & \textbf{83.77 $\pm$ 0.57} \\
\midrule
\multirow{3}{*}{CIFAR-100N} & \multirow{3}{*}{$\approx 0.4$}
& Acc. & 43.91 $\pm$ 0.87 & \textbf{47.60 $\pm$ 2.09} \\
& & F1 & 64.20 $\pm$ 1.05 & \textbf{65.89 $\pm$ 1.63} \\
& & BA & 69.72 $\pm$ 0.77 & \textbf{71.23 $\pm$ 1.40} \\
\bottomrule
\end{tabular}
}
\end{table}

\paragraph{ImageNet pre-trained weights} 
Often, in practice, models are initialized with  ImageNet pre-trained weights, which are widely regarded as strong general-purpose feature extractors~\citep{krizhevsky2012imagnet}. To assess the real-world relevance of our approach, 
we compare our SSL pre-training scheme directly against ImageNet weight initialization in \Cref{tab:imagenet_pretrain}.
At low corruption rates, both initialization strategies yield similar performance. However, as label noise increases, ImageNet pre-training offers little to no advantage over training from scratch, while SSL pre-training continues to provide substantial improvements. This highlights the benefit of domain-aligned self-supervised pre-training when working with noisy datasets particularly at high noise rates.

\begin{table}[!h]
\centering
\caption{Comparison of baseline, ImageNet pre-training, and SSL pre-training under increasing label noise on CIFAR-10.}
\label{tab:imagenet_pretrain}
\resizebox{0.95\columnwidth}{!}{
\begin{tabular}{c|ccc}
\toprule
\multirow{2}{*}{\textbf{$\eta$ }} & \multicolumn{3}{c}{\textbf{Accuracy (\%)}} \\
\cmidrule(lr){2-4}
 & Baseline & ImageNet & SimCLR (SSL) \\
\midrule
0.0 & 83.27 $\pm$ 1.06 & \textbf{89.43 $\pm$ 0.78} & 88.63 $\pm$ 1.20 \\
0.2 & 80.42 $\pm$ 1.09 & \textbf{87.25 $\pm$ 0.51} & 86.46 $\pm$ 1.03 \\
0.4 & 73.00 $\pm$ 2.19 & 81.88 $\pm$ 1.62 & \textbf{82.76 $\pm$ 0.66} \\
0.6 & 59.50 $\pm$ 1.74 & 74.02 $\pm$ 1.52 & \textbf{77.50 $\pm$ 0.68} \\
0.8 & 27.28 $\pm$ 2.00 & 33.67 $\pm$ 7.22 & \textbf{57.16 $\pm$ 2.74} \\
\bottomrule
\end{tabular}
}
\end{table}

\begin{figure}[h]
    \centering
    \includegraphics[width=\columnwidth]{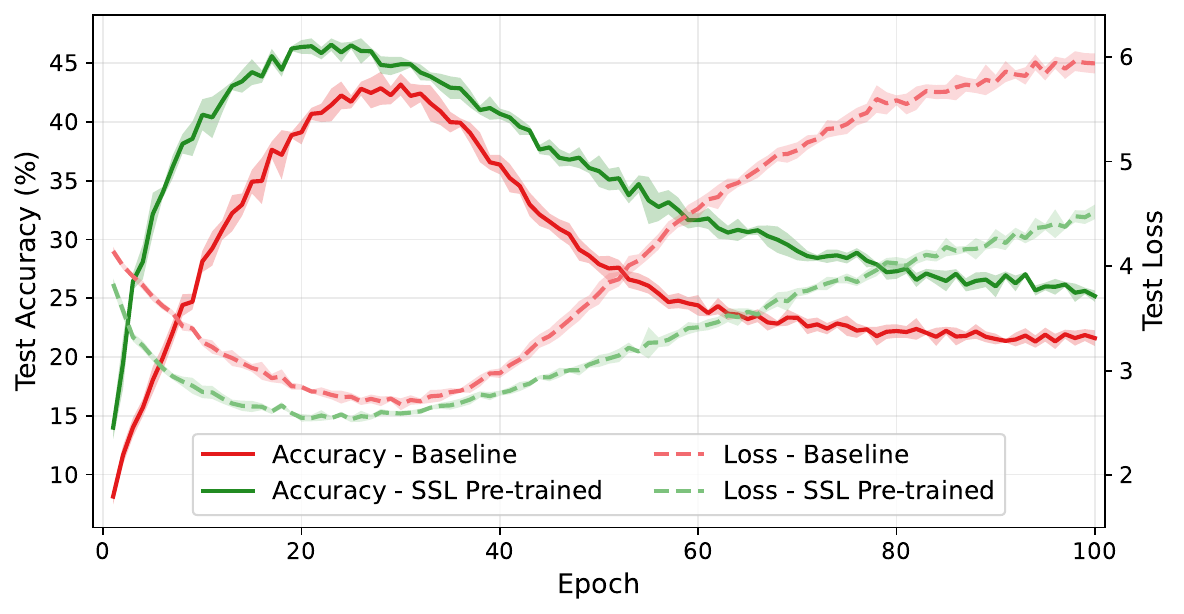}
    \caption{Comparison of Baseline and SSL-pretrained models over 100 supervised epochs on CIFAR-100 with $\eta=0.6$. 
    SSL pre-training (using SimCLR for 25 epochs) yields higher accuracy, slower overfitting, and reduced loss escalation under label noise.}
    \label{fig3}
\end{figure}

\paragraph{Increased Supervised Training} 
To test whether the benefits of SSL pre-training persist when models are allowed substantially more supervised training, we extend the fine-tuning stage from 10 to 100 epochs. As shown in \Cref{fig3}, averaged over five seeds, models trained from scratch take longer to reach peak performance and then rapidly overfit to the corrupted labels, causing a sharp decline in test accuracy and a pronounced rise in training loss.
In contrast, SSL pretrained models remain far more stable throughout training: they achieve higher accuracy, overfit more slowly, and exhibit a much weaker increase in loss during the overfitting phase. This behaviour indicates that the noise-resilient representations learned during pre-training continue to shield the model from memorizing corrupted labels, and even last for prolonged supervised training.
These results highlight that the gains from SSL pre-training are not merely due to increased total training duration, but from learning in-domain representations without the false signal of corrupt labels.

\begin{figure}[t]
    \centering
    \includegraphics[width=0.9\columnwidth]{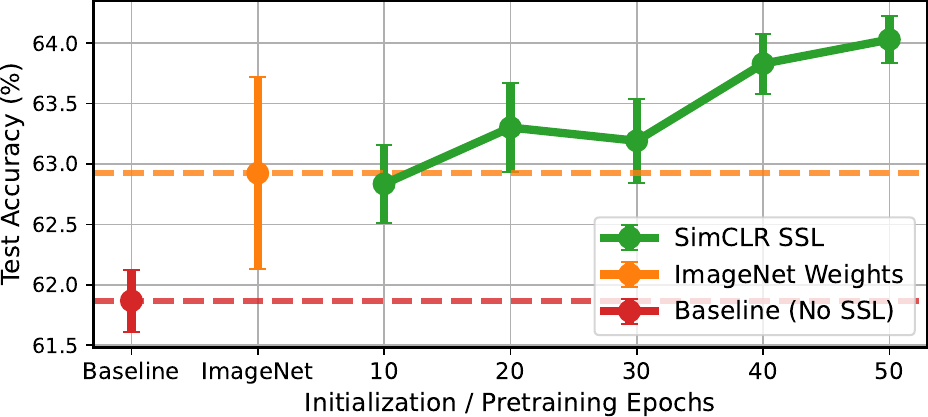}
    \caption{Test accuracy performance on Clothing-1M dataset with initialization from scratch, ImageNet weights, and in-domain SSL pre-trained weights using various durations of epochs with SimCLR on the training set. }
    \label{C1M_ssl}
\end{figure}

\paragraph{Large-scale Real-world Data} 
To evaluate whether the benefits of in-domain self-supervised pre-training extend beyond small-to-medium benchmark settings, we replicate our experimental pipeline on the Clothing-1M dataset~\citep{Xiao2015Clothing-1m}, which contains approximately 1 million images with inherent label noise arising from automatically collected online labels $(\eta \approx 0.4)$. We follow the same two-stage pipeline using a ResNet-50 backbone, with SimCLR pre-training on the full dataset followed by supervised fine-tuning until gains plateau (10 epochs).
As shown in \Cref{C1M_ssl}, in-domain SSL pre-training consistently improves test accuracy over both training from scratch and ImageNet initialization. Notably, with only 20 epochs of SSL pre-training, the model surpasses ImageNet initialization despite requiring no external labeled data.
This result highlights how in-domain SSL benefits from large scale datasets, where the pre-training leverages a larger amount of data. Additionally, unlike ImageNet pre-training, which introduces features learned from a different domain through transfer learning, SSL pre-training learns representations directly from the target distribution while remaining independent of corrupted labels.

\begin{figure*}[!ht]
    \centering
    \begin{subfigure}[b]{0.31\linewidth}
        \centering
        \includegraphics[width=\linewidth]{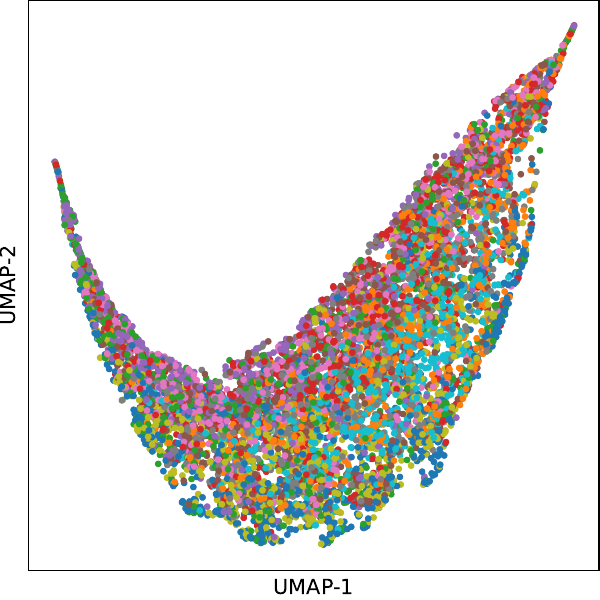}
        \caption{Random initialization}
    \end{subfigure}
    \hfill
    \begin{subfigure}[b]{0.31\linewidth}
        \centering
        \includegraphics[width=\linewidth]{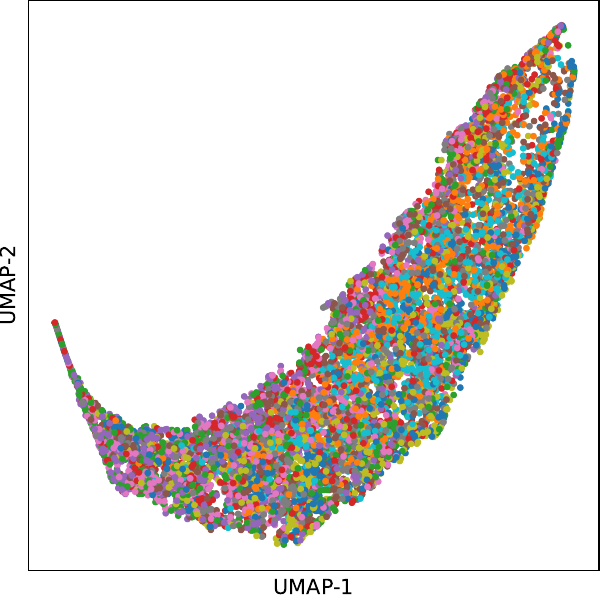}
        \caption{ImageNet initialization}
    \end{subfigure}
    \hfill
    \begin{subfigure}[b]{0.31\linewidth}
        \centering
        \includegraphics[width=\linewidth]{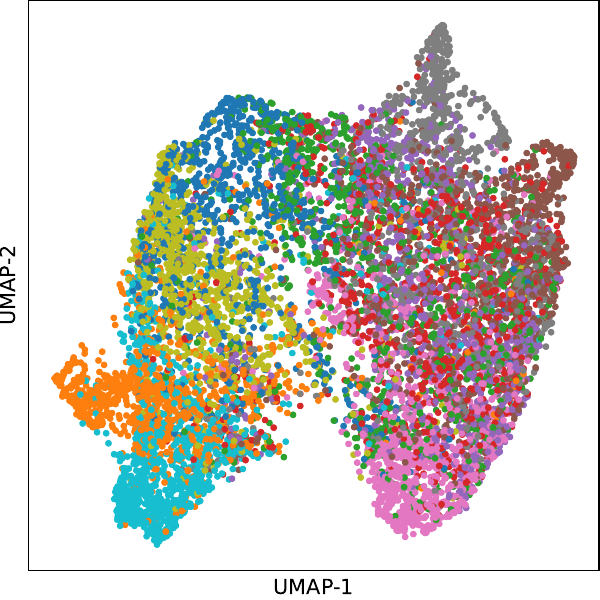}
        \caption{SimCLR SSL initialization}
    \end{subfigure}
    \caption{UMAP embeddings of features under different initialization strategies, prior to any supervised training on CIFAR-10 dataset. Only the SSL initialization using SimCLR for 50 epochs shows any class clustering in the latent space.}
    \label{fig:umap_init}
\end{figure*}

\paragraph{UMAP Analysis of Learned Representations}
To better understand how initialization strategies influence learned feature representations, we visualize embeddings using UMAP, prior to supervised training \citep{mcinnes2020umap}. Features are extracted from the encoder and projected into two dimensions to analyze class separability in latent space. As shown in \Cref{fig:umap_init}, random initialization produces unstructured embeddings with no discernible clustering. 
ImageNet initialization provides stronger general-purpose features, but these are learned from an external distribution and do not align with the target dataset, resulting in no class separation prior to fine-tuning.
In contrast, in-domain SSL pre-training produces class-aligned clusters before supervised training, indicating that meaningful semantic structure is learned directly from the target data. 
As a result, models achieve better downstream performance while relying less on potentially corrupted labels, as decision boundaries are already partially formed in the feature space.
We additionally provide UMAP visualizations after fine-tuning on CIFAR-10 under varying label noise in \Cref{fig:umap_grid} in Appendix. 
To quantify these observations, we compute silhouette scores (\Cref{tab:UMAP_Silhouette}), 
where higher values indicate more compact, well-separated clusters \citep{Rousseeuw1987silhouette, simpson2026clustering}.
Across all noise levels, SSL-pretrained models maintain 
higher silhouette scores compared to random initialization, and retain performance under high corruption ($\eta=0.6$), while ImageNet features degrade, as they rely more heavily on noisy supervision during fine-tuning. 
These results support our hypothesis that learning representations independent of labels reduces reliance on noisy supervision and improves robustness.

\begin{table}[ht]
\centering
\caption{Silhouette scores for UMAP with different initializations. Models fine-tuned under increasing noise on CIFAR-10. ImageNet and SSL improve latent class separation, but SSL outperforms ImageNet at higher noise rates.}
\label{tab:UMAP_Silhouette}
\resizebox{0.96\columnwidth}{!}{
\begin{tabular}{c|ccc}
\toprule
\multirow{2}{*}{\textbf{$\eta$}} & \multicolumn{3}{c}{\textbf{Initialization Method}} \\
\cmidrule(lr){2-4}
 & Baseline & ImageNet & SimCLR (SSL) \\
\midrule
0.0 & 0.214 $\pm$ 0.012 & \textbf{0.365 $\pm$ 0.008} & 0.327 $\pm$ 0.003 \\
0.2 & 0.265 $\pm$ 0.011 & \textbf{0.403 $\pm$ 0.014} & 0.376 $\pm$ 0.007 \\
0.4 & 0.180 $\pm$ 0.023 & 0.313 $\pm$ 0.031 & \textbf{0.339 $\pm$ 0.003} \\
0.6 & 0.076 $\pm$ 0.015 & 0.134 $\pm$ 0.084 & \textbf{0.265 $\pm$ 0.006} \\
\bottomrule
\end{tabular}
}
\end{table}

\begin{figure*}[!ht]
    \centering
    \begin{subfigure}[t]{0.44\textwidth}
        \centering
        \includegraphics[width=\linewidth]{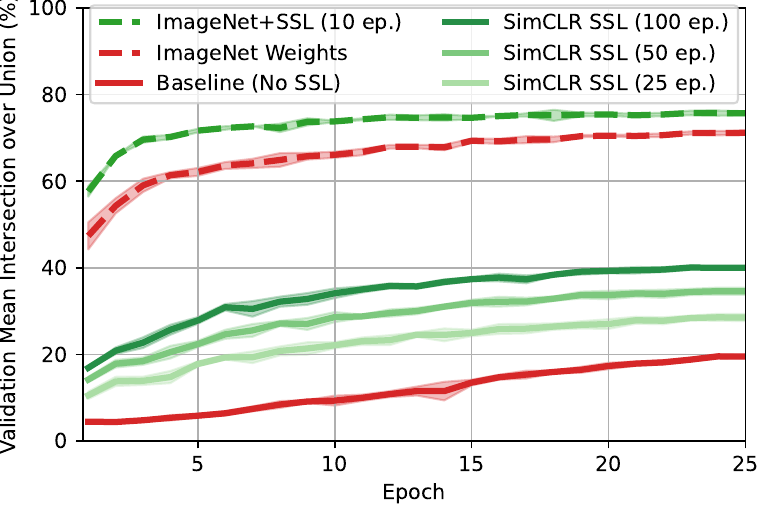}
        \caption{Performance at $\eta = 0.0$ over 25 epochs}
        \label{fig:voc_mIoU_noise-0.0}
    \end{subfigure}%
    ~~~
    \begin{subfigure}[t]{0.44\textwidth}
        \centering
        \includegraphics[width=\linewidth]{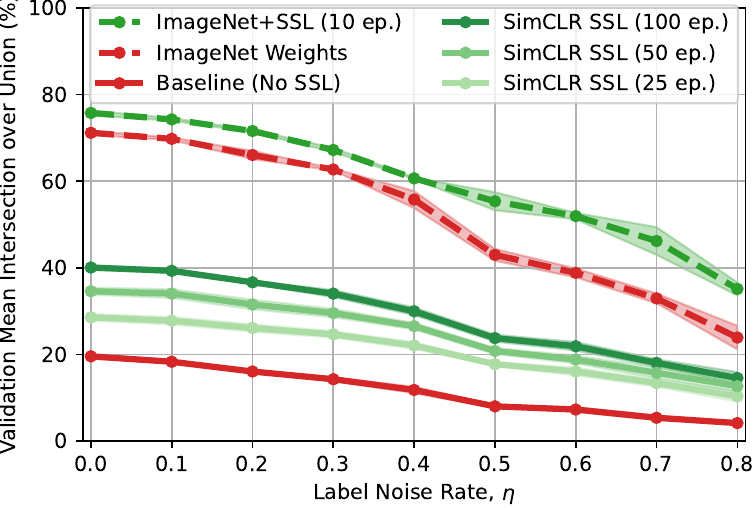}
        \caption{Peak performance for various $\eta$ values}
        \label{fig:voc_noise_sweep}
    \end{subfigure}
    \caption{Mean Intersection over Union during supervised training on VOC2012 dataset using various initialization methods. Results from training on clean labels shown in (\subref{fig:voc_mIoU_noise-0.0}), whereas (\subref{fig:voc_noise_sweep}) shows peak performance achieved by models for every \mbox{noise rate $\eta$.}}
    
    \label{fig:voc_mIoU}
\end{figure*}

\paragraph{Extension to Semantic Segmentation} 
Since self-supervised pre-training learns representations independently of the downstream task, the proposed pipeline is not limited to image-level classification. To evaluate its generality, we extend it to semantic segmentation using the PASCAL VOC 2012 dataset \citep{everingham2010pascal}.
As shown in \Cref{fig:voc_mIoU_noise-0.0}, SSL pre-training improves mean Intersection over Union (mIoU) performance compared to training from scratch. Due to the limited size of the segmentation training set (\textasciitilde 1.5k images), ImageNet initialization provides a strong baseline, significantly outperforming standalone SSL pre-training. This highlights a key limitation of in-domain SSL in low-data regimes.
However, combining both approaches yields the best performance. Initializing with ImageNet weights followed by in-domain SSL pre-training produces consistently higher mIoU than either method alone, with only 10 epochs of SSL required to reach peak performance. These results demonstrate that in-domain SSL pre-training is complementary to traditional transfer learning, and can be effectively applied beyond image classification to improve robustness in structured prediction tasks such as segmentation.  We additionally corrupt the training set by randomly reassigning individual segmentation masks to a different class label with a rate of $\eta$. Results in \Cref{fig:voc_noise_sweep} show that SSL pre-training consistently improves peak performance over baseline methods at all noise levels. With just 10 epochs of SSL pre-training using ImageNet initialization, peak performance is additionally improved over standard ImageNet initialization, and the performance gap increases at higher noise rates ($\eta \ge 0.5$).

\begin{table}[ht]
\centering
\caption{Comparison of Vision Transformer SSL methods 
on EuroSAT dataset under increasing synthetic label noise. Values report downstream clean-test accuracy (\%) after supervised fine-tuning for 10 epochs. The baseline trains the ViT from random initialization, while the DINOv2-weight baselines use externally pre-trained weights evaluated with either linear probing or full fine-tuning.}
\label{tab:vit_ssl}
\resizebox{\columnwidth}{!}{
\begin{tabular}{lccc}
\toprule
\textbf{Init. Method} & \textbf{$\eta=0.0$} & \textbf{$\eta=0.2$} & \textbf{$\eta=0.4$} \\
\midrule
Baseline & 90.02 $\pm$ 1.09 & 87.19 $\pm$ 0.56 & 83.19 $\pm$ 0.87 \\
VICReg & 95.28 $\pm$ 0.36 & 94.72 $\pm$ 0.32 & 93.61 $\pm$ 0.76 \\ 
SimMIM & 95.41 $\pm$ 0.87 & 94.27 $\pm$ 0.45 & 92.56 $\pm$ 0.87 \\
MAE & 95.71 $\pm$ 0.44 & 94.35 $\pm$ 0.52 & 92.20 $\pm$ 0.80 \\
iBOT & 96.92 $\pm$ 0.38 & 96.20 $\pm$ 0.54 & 95.46 $\pm$ 0.30 \\
DINO & 96.99 $\pm$ 0.20 & 96.32 $\pm$ 0.28 & 95.25 $\pm$ 0.26 \\
DINOv2 & 96.11 $\pm$ 0.22 & 95.13 $\pm$ 0.36 & 94.15 $\pm$ 0.42 \\
MSN & \textbf{97.58 $\pm$ 0.19} & \textbf{97.12 $\pm$ 0.16} & \textbf{96.22 $\pm$ 0.40} \\
\midrule
\multicolumn{4}{l}{\textbf{DINOv2 Weights}} \\
\cmidrule(l){1-4}
Linear Probed & 94.44 $\pm$ 0.27 & 93.60 $\pm$ 0.47 & 92.56 $\pm$ 0.28 \\
Fine-tuned & 96.78 $\pm$ 0.35 & 95.52 $\pm$ 0.89 & 92.94 $\pm$ 0.67 \\
\bottomrule
\end{tabular}
}
\end{table}

\begin{figure}[h]
    \centering
    \includegraphics[width=\columnwidth]{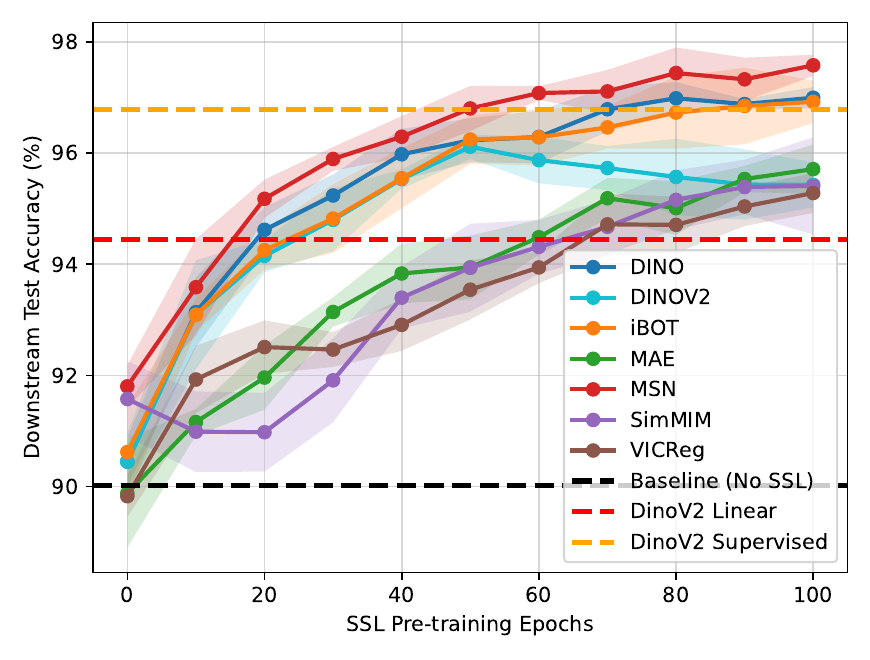}
    \caption{Downstream EuroSAT test accuracy after different durations of in-domain SSL pre-training, and subsequent finetuning ($\eta=0$), shown as solid lines. The dashed black line denotes the no-SSL baseline, while the red and orange dashed lines denote DINOv2-weight baselines evaluated with linear probing and full fine-tuning, respectively. In-domain ViT SSL pre-training consistently improves over the baseline and linear probing on DinoV2 weights, with several methods outperforming finetuning on DinoV2 weights.}
\label{fig:vit_ssl}
\end{figure}

\paragraph{Extension to Vision Transformer Architectures} 
To evaluate whether the proposed in-domain SSL pre-training strategy generalizes beyond convolutional architectures, we additionally perform experiments using Vision Transformers. This setting is important because modern SSL methods are increasingly designed around transformer backbones, and strong off-the-shelf ViT representations such as DINOv2 provide a demanding comparison point. We therefore evaluate several ViT-compatible SSL objectives on the EuroSAT dataset \citep{helber2019eurosat}. Each method is pre-trained for 100 epochs on the target training images without labels, then fine-tuned for 10 epochs using supervised training on the test set under varying levels of synthetic label noise. This mirrors the main experimental pipeline, but replaces the ResNet-based backbone with a ViT-S/16 architecture.  
The results in \Cref{tab:vit_ssl} show that in-domain ViT SSL pre-training consistently improves downstream classification accuracy over training the same ViT from random initialization, with significant improvements on datasets with higher amounts of label noise. This trend is consistent with the ResNet-based experiments, suggesting that the benefits of SSL pre-training are not tied to a particular architecture family. Instead, the main advantage appears to come from learning target-domain visual structure before exposing the model to supervised labels.
Although DINOv2 weights remain a strong general-purpose initialization, the in-domain SSL models are competitive with this reference point and, for several methods, surpass it on \mbox{EuroSAT} after sufficient in-domain pre-training (\Cref{fig:vit_ssl}). This suggests that target-domain SSL can efficiently
recover much of the benefit of large-scale pre-training, and in some cases improve upon it when the downstream domain differs from the original pre-training distribution, without requiring large amounts of data or training time.

\section{Discussion and Conclusion}

Our results demonstrate that in-domain self-supervised pre-training consistently improves downstream performance across diverse datasets, architectures, and downstream tasks, under both clean and noisy supervision. The gains are not tied to a particular SSL objective, but instead arise from learning in-domain representations before supervised optimization. Because this representation learning is independent of potentially corrupted labels, fine-tuning begins from a feature space that already captures the underlying semantic structure of the data, leading to higher classification accuracy, improved label-error detection, and reduced memorization of incorrect annotations. The advantage of SSL becomes increasingly pronounced as label quality deteriorates, suggesting that robust representations are especially valuable when reliable supervision is limited.

UMAP visualizations, silhouette analysis, and improved label-error detection provide complementary evidence that SSL learns compact, well-separated feature representations that remain robust under noisy supervision. While generic pre-trained weights such as ImageNet and DINOv2 provide strong initializations, particularly in low-data settings, our experiments show that combining transfer learning with in-domain SSL outperforms either strategy alone. Although the additional pre-training stage introduces extra computation, most gains are achieved within relatively few SSL epochs, making the approach practical without requiring modifications to the downstream supervised training pipeline.

Collectively, these results suggest robustness arises not from any individual SSL algorithm, but from learning task-specific representations before fitting to noisy labels during supervised optimization. By separating representation learning from corrupted labels, in-domain SSL provides a simple, architecture-agnostic, and broadly applicable strategy for improving robustness across a wide range of learning settings. Future work will investigate larger foundation models and additional modalities to further understand the limits of this approach.

\bibliography{aaai2027}


\appendix

\section{Technical Appendices and Supplementary Material}
\label{sec:appendix}

\begin{figure*}[ht]
    \centering

    \begin{subfigure}[b]{0.31\linewidth}
        \centering
        \includegraphics[width=\linewidth]{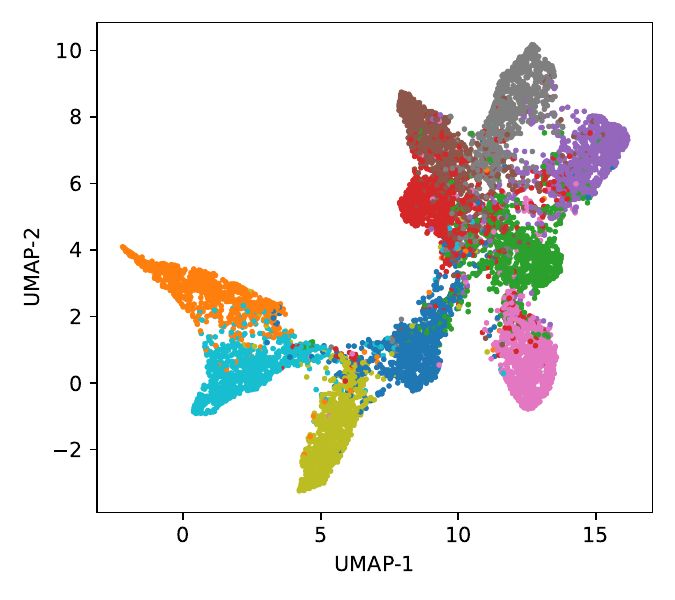}
        \caption{Random, $\eta=0.0$}
    \end{subfigure}
    \hfill
    \begin{subfigure}[b]{0.31\linewidth}
        \centering
        \includegraphics[width=\linewidth]{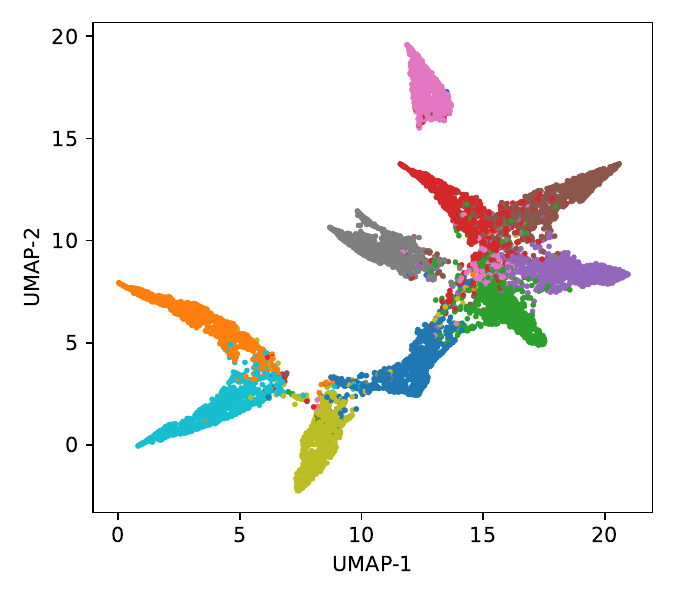}
        \caption{ImageNet, $\eta=0.0$}
    \end{subfigure}
    \hfill
    \begin{subfigure}[b]{0.31\linewidth}
        \centering
        \includegraphics[width=\linewidth]{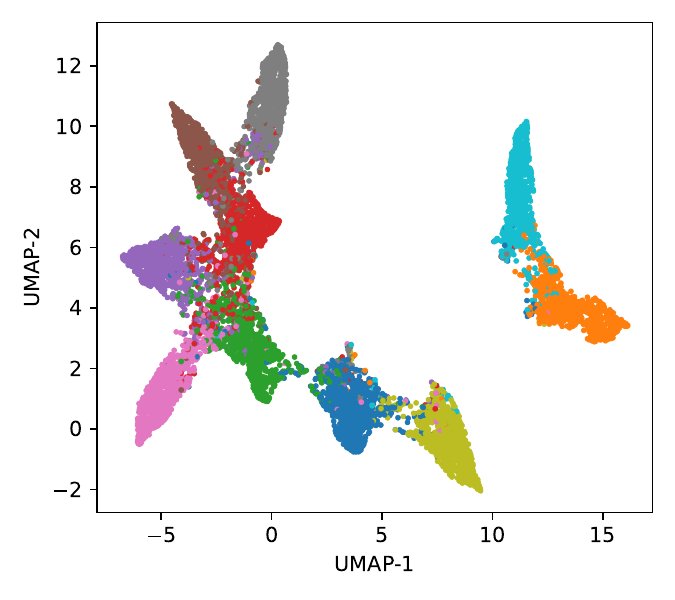}
        \caption{SSL, $\eta=0.0$}
    \end{subfigure}

    \vspace{2.0em}

    \begin{subfigure}[b]{0.31\linewidth}
        \centering
        \includegraphics[width=\linewidth]{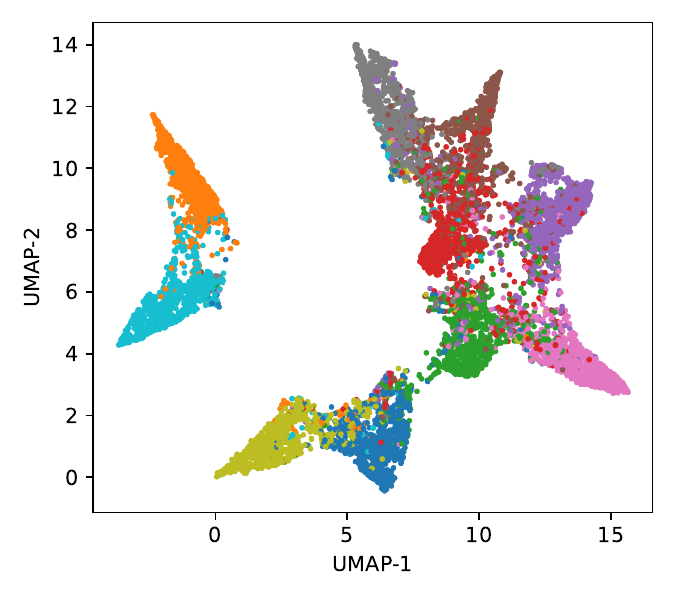}
        \caption{Random, $\eta=0.2$}
    \end{subfigure}
    \hfill
    \begin{subfigure}[b]{0.31\linewidth}
        \centering
        \includegraphics[width=\linewidth]{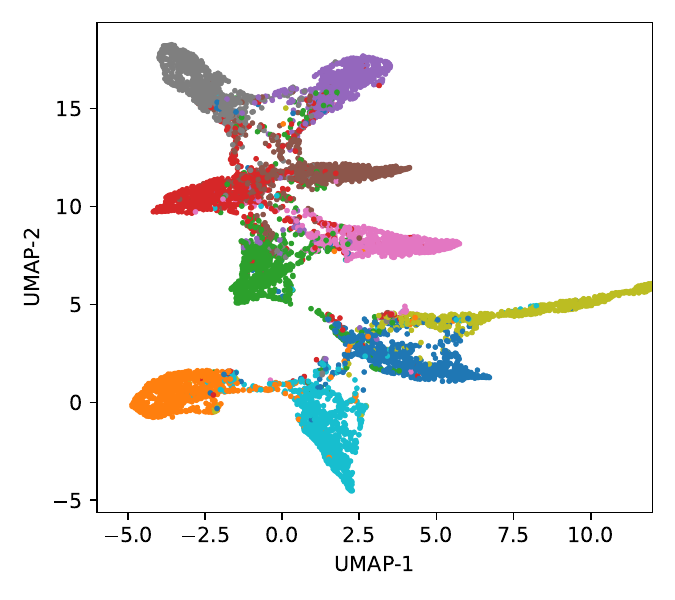}
        \caption{ImageNet, $\eta=0.2$}
    \end{subfigure}
    \hfill
    \begin{subfigure}[b]{0.31\linewidth}
        \centering
        \includegraphics[width=\linewidth]{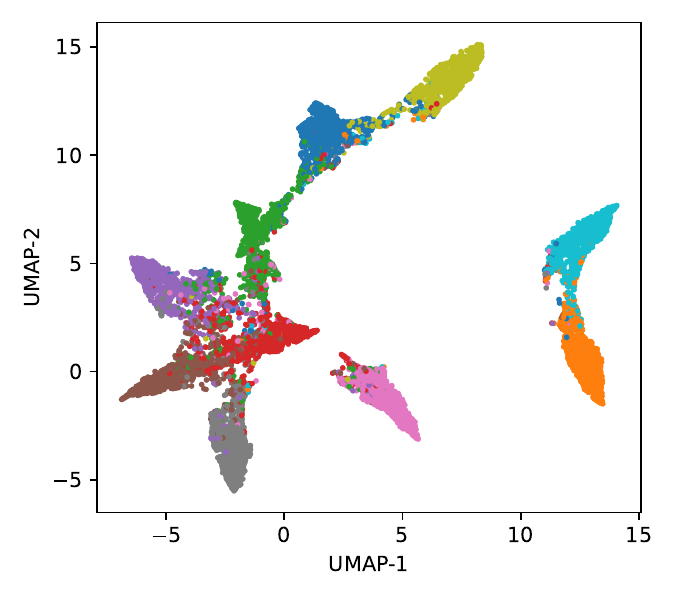}
        \caption{SSL, $\eta=0.2$}
    \end{subfigure}

    \vspace{2.0em}

    \begin{subfigure}[b]{0.31\linewidth}
        \centering
        \includegraphics[width=\linewidth]{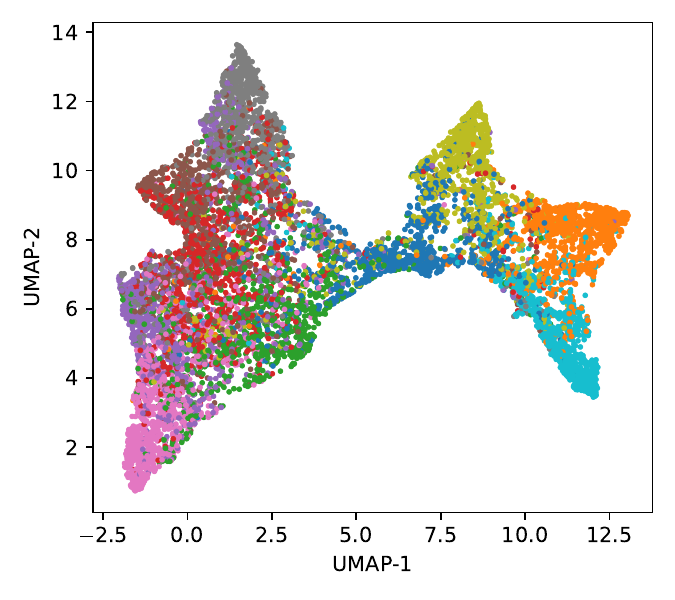}
        \caption{Random, $\eta=0.6$}
    \end{subfigure}
    \hfill
    \begin{subfigure}[b]{0.31\linewidth}
        \centering
        \includegraphics[width=\linewidth]{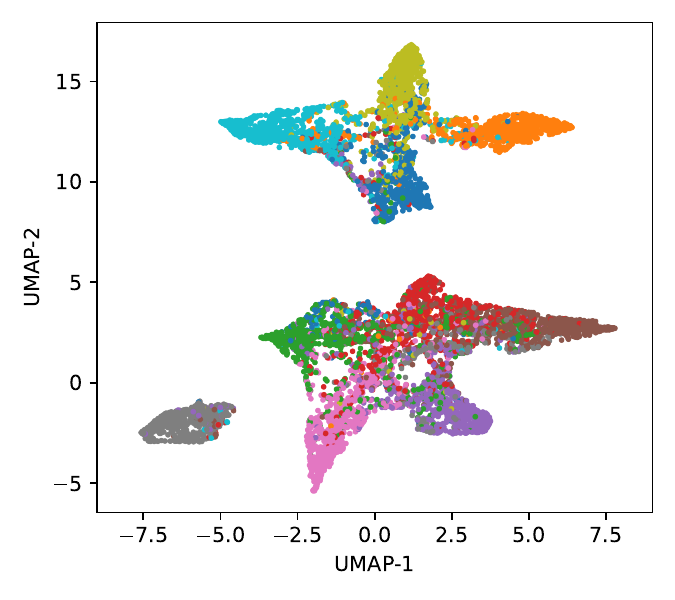}
        \caption{ImageNet, $\eta=0.6$}
    \end{subfigure}
    \hfill
    \begin{subfigure}[b]{0.31\linewidth}
        \centering
        \includegraphics[width=\linewidth]{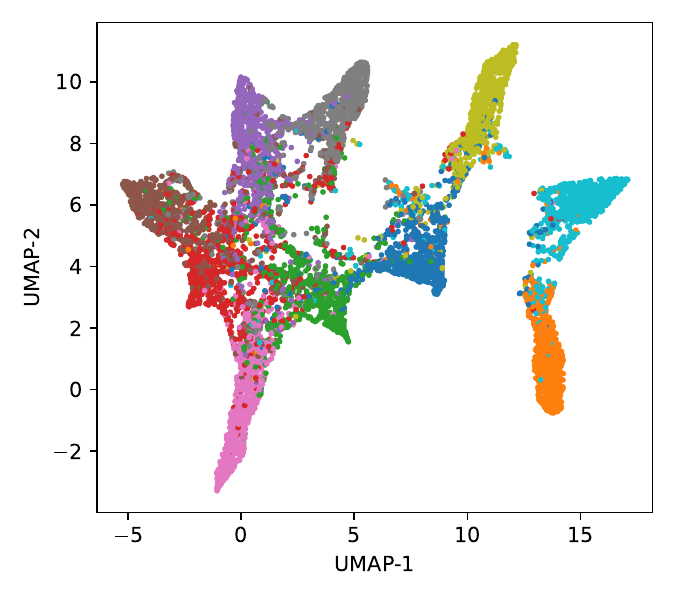}
        \caption{SSL, $\eta=0.6$}
    \end{subfigure}

    \caption{
    UMAP embeddings corresponding to a fixed noise level $\eta$ in each row, while columns compare Random, ImageNet, and SSL initializations after fine-tuning for 10 epochs on CIFAR-10 dataset.
    Pre-training on the train set for 50 epochs using SimCLR SSL maintains more distinct class clustering over the baseline as noise increases.
    }
    \label{fig:umap_grid}
\end{figure*}

\paragraph{Reproducibility details}
All experiments were implemented in Python 3.11.5. Additional software dependencies and package versions are provided in the \texttt{requirements.txt} file included with our code release. The repository contains all code required for data preprocessing, self-supervised pre-training, supervised training, evaluation, label-error detection, and figure generation, and is available through our repository.


\paragraph{Compute Resource Requirements}
All experiments were conducted using NVIDIA A100 MIG GPU partitions. CIFAR-10 and CIFAR-100 experiments used an A100 1g.5gb partition (5~GB GPU memory), while the Clothing1M, PASCAL VOC, and Vision Transformer experiments used an A100 3g.20gb partition (20~GB GPU memory). Compute nodes were equipped with dual AMD EPYC 7532 (Zen~2) processors (64 CPU cores total, 2.40\,GHz).

Self-supervised pre-training on CIFAR required approximately 4~hours for a 100-epoch encoder, with shorter pre-training schedules scaling approximately linearly with the number of epochs. Standard CIFAR supervised fine-tuning (10 epochs) required approximately 10 minutes per noise rate and random seed, while the extended 100-epoch fine-tuning experiments required approximately 2~hours per configuration. Clothing1M pre-training required approximately 4~hours per seed, while supervised fine-tuning required approximately 10~hours per seed. PASCAL VOC SimCLR pre-training required approximately 5~hours per seed, and each supervised segmentation experiment required approximately 20 minutes. Vision Transformer experiments required an average of approximately 4~hours per seed for a single SSL method.
Overall, reproducing the complete experimental study, including all self-supervised methods, baseline comparisons, noise-rate sweeps, pre-training duration studies, and five random seeds, requires approximately 4,000 GPU-hours. An additional $\sim$500 GPU-hours were used during method development and \mbox{preliminary hyperparameter exploration.}

\paragraph{Implementation and Hyperparameter Details}
Unless otherwise stated, all results are averaged over five random seeds (\(\{1,2,3,4,5\}\)) and reported as mean \(\pm\) one standard deviation. Python, NumPy, and PyTorch random seeds were fixed at the start of each run with deterministic cuDNN settings enabled. Final optimization hyperparameters are summarized in \Cref{tab:final-hparams}.

CIFAR experiments used a CIFAR-adapted ResNet-18 with a \(3\times3\), stride-one input convolution and no initial max-pooling. Supervised training used standard random cropping and horizontal flipping. SimCLR employed the augmentation pipeline of random resized crops, horizontal flipping, color jitter, random grayscale, and Gaussian blur.

Clothing-1M experiments used a ResNet-50 backbone. Images were resized to 256 pixels, randomly cropped to \(224\times224\), randomly flipped during training, and normalized using ImageNet statistics.

PASCAL VOC 2012 experiments used DeepLabV3 with a ResNet-50 backbone. Images and segmentation masks were resized to 320 pixels before random \(256\times256\) crops and horizontal flipping. Segmentation performance was evaluated using mean Intersection-over-Union (mIoU), and synthetic label noise was generated by randomly relabeling connected components within segmentation masks.

EuroSAT experiments used ViT-S/16 at \(224\times224\) resolution and evaluated DINO, DINOv2, iBOT, MAE, MSN, SimMIM, and VICReg. Experiments used 100 epochs of SSL pre-training and noise rates of \(\eta\in\{0.0,0.2,0.4\}\).

\begin{table*}[h]
\centering
\caption{Experiment hyper-parameter settings.}
\label{tab:final-hparams}
\small
\resizebox{\textwidth}{!}{
\begin{tabular}{llllllllll}
\toprule
Dataset & Backbone & Method & SSL epochs & Sup. epochs &
SSL LR / WD & SSL batch & Sup. LR / WD & Sup. batch & Optimizer \\
\midrule
CIFAR-10/100 & ResNet-18 & SimCLR & 100 & 10 &
\(10^{-3}/10^{-4}\) & 256 & \(10^{-3}/10^{-4}\) & 256 & Adam \\
CIFAR-10N/100N & ResNet-18 & SimCLR & 50 & 10 &
\(10^{-3}/10^{-4}\) & 256 & \(10^{-3}/10^{-4}\) & 256 & Adam \\
Clothing-1M & ResNet-50 & SimCLR & 50 &
15 & \(3{\times}10^{-4}/10^{-4}\) & 128 &
\(10^{-3}/10^{-4}\) & 128 & Adam \\
VOC2012 & deeplabv3-R50 & SimCLR & 25/50/100 &
25 & \(3{\times}10^{-4}/10^{-4}\) & 64 &
\(10^{-4}/10^{-4}\) & 16 & AdamW \\
EuroSAT & ViT-S/16 & ViT SSL & 100 & 10 &
\(10^{-4}/0.05\) & 32 &
\(10^{-4}/10^{-4}\) & 64 & AdamW \\
\bottomrule
\end{tabular}
}
\end{table*}

\end{document}